\documentclass[preprint,10pt,5p,twocolumn]{elsarticle}
\usepackage{graphicx}
\usepackage{amssymb}
\usepackage{booktabs}
\usepackage{longtable,tabu,supertabular}
\usepackage{float}
\restylefloat{table}
\restylefloat{figure}
\usepackage{hyperref}

\journal{Aquaculture Engineering}

\usepackage{color}			

\begin{document}

\begin{frontmatter}

\title{Decision Support Systems in Fisheries and Aquaculture:\\A
  systematic review}

\author[label2,label3]{Bj{\o}rn Magnus Mathisen\corref{cor1}}
\ead{bjornmagnus.mathisen@sintef.no}
\address[label2]{SINTEF ICT\\Trondheim, Norway}
\address[label3]{Department of Computer and Information Science,\\ Norwegian University of Science
and Technology,\\Trondheim, Norway}
\address[label1]{SINTEF Nord\\Troms{\o}, Norway}

\author[label2,label1]{Peter Haro}

\cortext[cor1]{Corresponding author}

\ead{peter.haro@sintef.no}

\author[label1]{B{\aa}rd Hanssen}
\ead{bard.hanssen@sintef.no}

\author[label1]{Sara Bj\"{o}rk}
\ead{sara.bjork@sintef.no}

\author[label2]{St{\aa}le Walderhaug}
\ead{stale.walderhaug@sintef.no}

\begin{abstract}
  Decision support systems help decision makers make better decisions in the
  face of complex decision problems (e.g. investment or policy decisions).
  Fisheries and Aquaculture is a domain where decision makers face such
  decisions since they involve factors from many different scientific fields. No
  systematic overview of literature describing decision support systems and
  their application in fisheries and aquaculture has been conducted. This paper
  summarizes scientific literature that describes decision support systems
  applied to the domain of Fisheries and Aquaculture. We use an established
  systematic mapping survey method to conduct our literature mapping. Our
  research questions are: What decision support systems for fisheries and
  aquaculture exists? What are the most investigated fishery and aquaculture
  decision support systems topics and how have these changed over time? Do any
  current DSS for fisheries provide real-time analytics? Do DSSes in Fisheries
  and Aquaculture build their models using machine learning done on captured and
  grounded data? The paper then detail how we employ the systematic mapping
  method in answering these questions. This results in 27 papers being
  identified as relevant and gives an exposition on the primary methods
  concluded in the study for designing a decision support system. We provide an
  analysis of the research done in the studies collected. We discovered that
  most literature does not consider multiple aspects for multiple stakeholders
  in their work. In addition we observed that little or no work has been done
  with real-time analysis in these decision support systems.
\end{abstract}


\end{frontmatter}

\section{Introduction}
\label{sec1}

Decision makers within the Fishery and Aquaculture\footnote{For this study we
  define ``aquaculture'' as the area dealing with cultivation and farming of
  aquatic biomass, whereas ``fisheries'' refer to traditional fishing e.g. trawl
  and line vessels.} Industry are facing complex decision problems every day.
Where to trawl with the boat tomorrow? Where do I apply to build my new
aquaculture installation? These are just two examples, but answering these
questions relies on knowledge from many different fields of science and
heterogeneous data from many different sources. Answering \textit{where
  should the boat trawl tomorrow}? relies on the fields of meteorology, fish
biology, economics, ocean modeling and more. In addition it could require data
from previous trawls (location, time and amount), weather forecast, ocean model
output, fish stock models and previous logs of fuel consumption. To come to a
decision in the face of all these variables and knowledge is very demanding, but
doing it successfully can help decision makers optimize the operation of the
business.

Systems designed for enabling decision makers to make informed
decisions in the face of a complex problem are called
Decision Support Systems (DSS). DSSs tries to combine domain and
technical knowledge and package it in a way that can be of practical
use for non-scientists\cite{lannan1993user}.

Decision support systems (DSS) stems from the
1960s~\cite{power2008decision}, and has been applied to a multitude of
domains. Although the taxonomy and general process of creating, using
and maintaining DSSs are well documented both in case studies and
research, the literature provides little information regarding
empirical assessments of its effectiveness in particular domains.
We could only find a few systematic literature review of decision support
systems research. However these were exclusively within the domain clinical
decision support systems within medicine has been subject to systematic reviews
targeting effects\cite{garg2005effects,hunt1998effects} and how to improve such
systems\cite{kawamoto2005improving}. More closely related to research on DSS in
fishery and aquaculture is research on spatial DSS, which has been studied in
non-systematic surveys\cite{crossland1995spatial,malczewski2006gis}

The primary research hypothesis of this paper postulates there is
little empirical knowledge of the effectiveness of decision support
based system in the fishery and aquaculture domain. In addition, it is
believed that there is no single system that combines both their
respective requirements to maximize economic efficiency,
sustainability, produce optimal vessel-scheduling and reduce
environmental footprints, based on multiple data sources.
Therefore
this study aims to aggregate primary empirical studies in an objective
manner to refute or support the primary hypothesis.

\textbf{Context:} There are little to no decision support system for
fisheries and aquaculture that supports a variety of fish
species. Neither does there exist a fully integrated information
system with near real-time advanced analysis models for fisheries.

\textbf{Objectives:} To conduct a mapping study to survey existing
research on DSSs in order to identify useful approaches and clarify
needs for further research.  Method: A systematic mapping study of the
available literature following the best practice methods laid out by previous
practitioners\cite{kitchenham2007}.

\textbf{Results:} 27 papers have been identified by topic, system
classification, and relevance for the fishery and aquaculture
domain. We found that fishery- and aquaculture decision support
systems rarely evaluate their system empirically, which indicates that
additional investigation, empirical and practical, should be
performed. In addition, the study found no DSS contradictory to the
context of this study; however it identifies key methodology and
insights for future usage in fishery- and aquaculture DSSs.

\textbf{Conclusions:} The majority of fishery- and aquaculture
decision support systems published over the last 25 years focus on
singular topics, and these systems does not provide multiple aspects
for multiple stakeholders on the consideration of multiple factors. We
observed empirical evaluation and real-time data analytics to be
virtually non-existent in the problem domain.

\section{Method}
\label{conduct}
To gather data on the state of decision support system research within the
domain of Fisheries and Aquaculture we apply a systematic literature review.
This way we comply to a well known and defined method, providing reproducability
and rigor while at the same time acquiring knowledge about the field and
answering our research questions.

This mapping study has been conducted in compliance with a pre-defined
protocol created for this study to reduce the possibility of
researcher bias~\cite{kitchenham2007} . The review protocol is an essential
component for providing context and domain classification, and a
protocol must be developed separately for each mapping study in order
to define the main guidelines for conducting systematic mapping
studies.  Both Kitchenham \cite{kitchenham2007} and Budgen
et al.~\cite{budgen2008using} states that the research questions in
mapping studies are likely to be broader than in traditional
Systematic Literature Reviews (SLR), to adequately address the wider
scope of the study. Kitchenham~\cite{kitchenham2007} also states
that mapping studies will likely return a very large number of studies
which in turn will give a much broader coverage than the outcome of
the SLR. On the basis of the preceding statements, a systematic
mapping study was selected as the method for achieving a broad
resolution on the research questions as opposed to an SLR.

The following four research questions (RQs) were formulated in order
to characterize the field of DSS within fishery and aquaculture:

\begin{itemize}
\item What decision support systems for fisheries and aquaculture
  exists? (RQ1) 
\item What are the most investigated fishery and
  aquaculture DSS topics and how have these changed over time? (RQ2) 
\item Do any current DSS for fisheries and aquaculture use real-time analytics?
  (RQ3)
\item Do DSSes in Fisheries and Aquaculture build their models
  using machine learning done on captured and grounded data? (RQ4)
\end{itemize}

\subsection{Rigor of Study}
\label{validity}
The study has been conducted by three of the authors in collaboration, with two
of the authors acting as a reviewers of the process. The search process of the
mapping study has been executed analogous to traditional SLR studies as similar
processes for searching are explicitly defined in the research protocol and
reported as part of the outcome~\cite{budgen2008using}. In compliance with
Kitchenham~\cite{kitchenham2007}, researchers should specify their rationale for
the use of electronic or manual searches or a combination of both. Although most
text books emphasize the use of electronic search procedures, they are not
usually sufficient by themselves, and some researchers strongly advocate the use
of manual searches (e.g.~\cite{kitchenham2007}). The results presented in this
study is a combination of both techniques, and is justified because the field of
maritime DSS with real-time analytics has not been researched sufficiently to
aggregate enough results through only automatic searches. The following sources
were used for this study:

\begin{itemize}
\item IEEE Explore 
\item CM Digital Library
\item Google Scholar
\item Citeseer library
\item Springer
\item Ei Compendex
\end{itemize}

These sources were selected because they are among the most important
repositories for acquiring data in computer science and collectively
they addressed the main digital libraries deemed appropriate for this
study. No researchers were contacted directly in this survey.  The
results retrieved from executing a search on the various data sources
were either dismissed or accepted into the primary study selection
process, based on inclusion- and exclusion criteria. The inclusion-
and exclusion criteria are used to exclude papers that are not
relevant to answer the research questions, and are one of the
activities in a mapping study near identical to
SLR~\cite{petticrew2008systematic}.  The study used the following
inclusion criteria:

\begin{enumerate}
\item Only studies written in English or Norwegian 
\item Studies
  noting or referencing any of the subjects described in the research
  question in their title or abstract 
\item Studies had to be published after 1990 \footnote{And not after the first
    half of 2015 as this is when this study finished.}
\item Studies had no restriction on geographical placement
\end{enumerate}
  And the following exclusion criteria:
\begin{enumerate}
\item Duplicate results found in
  another search engine. 
\item Analogous studies reporting similar
  results, only the most complete study was considered.
\item Inaccessible studies or books.
\item Literature that only was
  available in the form of PowerPoint presentations or abstracts.
\end{enumerate}

Additionally, a fifth exclusion criterion was added in order to tune
the search in order to exclude publications that described lake or
pond aquaculture or fisheries in such a way that the results would not be
applicable to offshore based fisheries and aquaculture:

\begin{enumerate}
\setcounter{enumi}{4}
\item No studies exclusively describing DSSs for fishery and/or aquaculture in
  lakes or ponds were considered.
\end{enumerate}

Finally the studies not contributing to answering any of the research questions
posed by this our study was excluded.

\begin{enumerate}
\setcounter{enumi}{5}
\item Only studies contributing to answering our research questions were considered.
\end{enumerate}

No quality assurance or assessment was performed during the search
phase, simply to achieve maximum coverage.  In agreement with our
review protocol, search terms were used for identifying relevant
papers in the field of fishery- and aquaculture DSS. We derived our
process for synthesizing the query strings in our review protocol from
Kitchenham et al.~\cite{kitchenham2007cross,kitchenham2007}. The search terms were
selected with trial from a candidate set, which were populated by
deriving terms from the research questions. The Boolean ``AND'' was used
to link keywords from different populations in the search strings, the
Boolean OR was also used to incorporate alternative spellings. We
ended up with the following final search strings:

\begin{itemize}
\item "Decision support system" AND "empirical evidence" (Q1)
\item Decision support system fisheries (Q2) 
\item Decision support system
  aquaculture (Q3) 
\item "Decision support system" AND "fisher*" AND
  (real-time OR realtime OR real time) (Q4) 
\item "(multi-criteria OR multicriteria OR multi criteria) decision
  making" AND "fishery management" (Q5)
\end{itemize}

It is important to note that we chose to not include ``operator
support''/``operational support'' that can be considered a weak synonym of DSS.
However operational support does not share main motivation/stakeholder as DSS
and focuses more on operational aspects. 

We used EndNote X7.3.1 as our reference manager, and to synthesize our
findings, we exported the result set to Microsoft Excel utilizing our
separate evaluations to categorize and record details of each
paper.

\section{Results}
\label{sec:outcome}
The method applied in order to identify relevant studies can be divided into
three discrete steps, where the first stage was applying the search query on the
data sources. The firs try resulted in 510681 hits. As a result the search
queries were refined into the query strings listed in the previous section.

After applying the refined queries across all sources it returned a total of 146
results. More specifically all five query strings where applied to all of the
search engines. The five results of from each source was collated, sorted then
compared to every other source. The resulting 146 documents where the greatest
common denominator of these source. After
applying the inclusion- and exclusion criteria, the sum of both manual and
automatic searches totaled 70 documents.

\subsection{Reading and selective filtering}
\label{sec:method}
The data-set containing 70 articles was separately evaluated. Each evaluator was
required to compile a list containing all papers complemented by a short
description of the paper and its contribution to the research questions this
study aims to resolve. Studies not contributing to answering any of the research
questions, according to the evaluators, where excluded. The separate evaluations
were discussed and compared in order to properly evaluate whether or not the
paper should be used in the study. In order to determine the contribution of the
articles a majority decision based on the compiled lists had to be reached, to
avoid any particular research bias. Juxtaposing the papers resulted in an
additional exclusion of 43 papers.

\begin{figure}[H]
  \centering
    \includegraphics[width=\columnwidth]{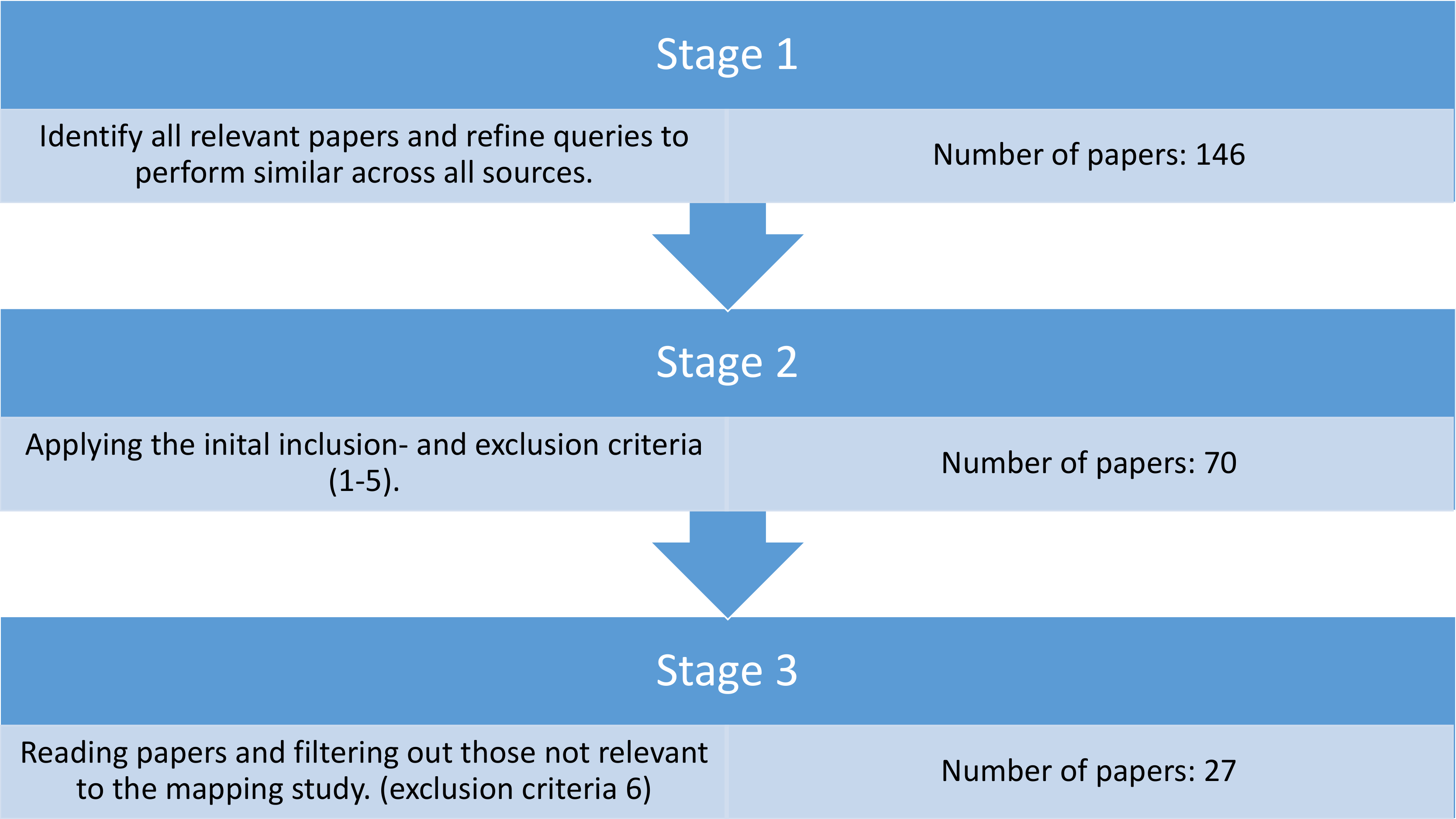}
    \caption{The different stages of the review. Stage }
  \label{fig:method}
\end{figure}

Figure~\ref{fig:method} summarizes how we implemented the selection process and
presents the number of papers that were identified in each step of the search
process. Appendix~\ref{sec:selectedstudies} presents a complete list of the 27
selected studies, numbered from SID-1 to SID-27.

\section{Analysis}
This analysis is heavily based on the research questions in the
mapping study and is primarily confined into two issues: what decision
making systems from fisheries and aquaculture exists, and which
empirical results does these systems provide, especially regarding
real-time analytics.

\subsection{Classification of selected studies and results}
\label{sec:classification}
\begin{figure}[H]
  \centering
    \includegraphics[width=\columnwidth]{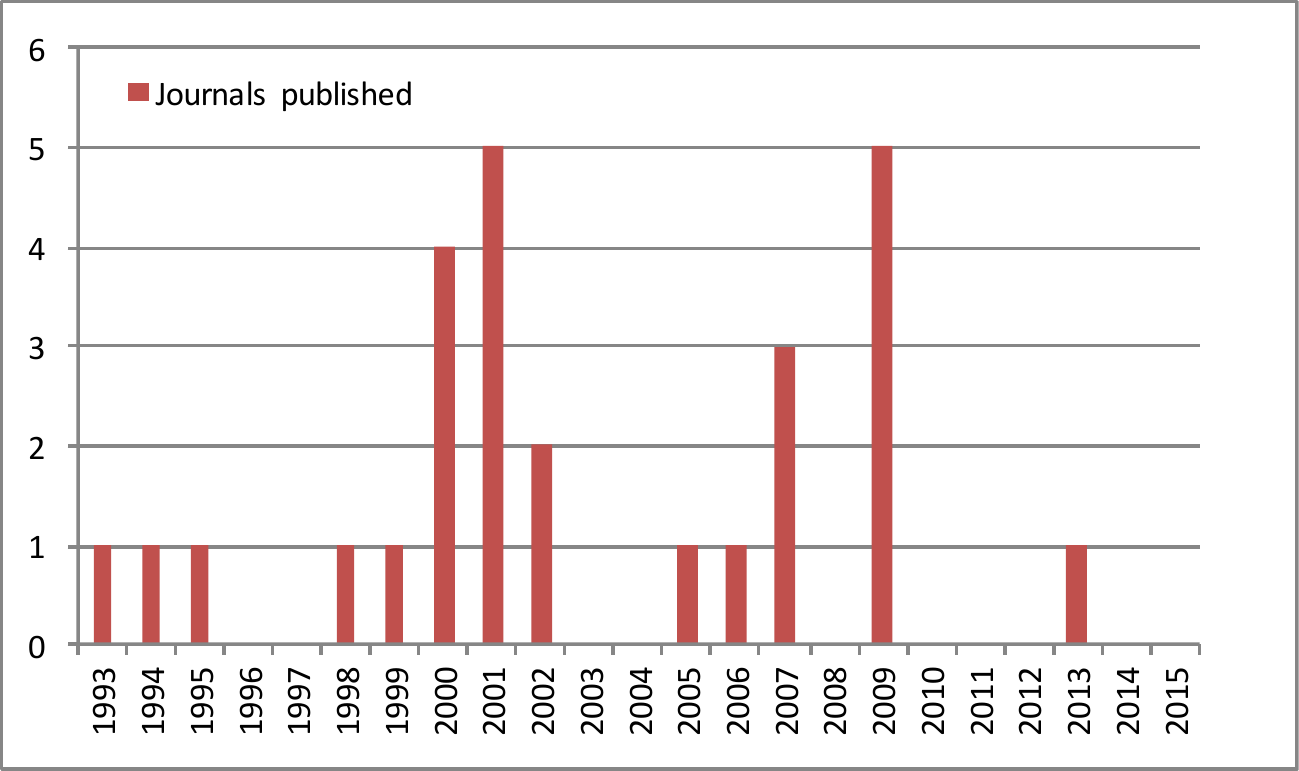}
    \caption{Publications published on DSS in fishery and aquaculture}
  \label{fig:journal}
\end{figure}

We present our findings in the form of a qualitative synthesis,
close to what Kitchenham~\cite{kitchenham2007} describes as a
line of argument synthesis as this study tries to infer as much domain
knowledge as possible. Consequential to the research questions and future work is
the research regarding fishery and aquaculture DSS systems; we
therefore start by presenting the number of journals found during
the search phase, sorted by publishing year. It should be noted that
the numbers presented in Figure~\ref{fig:journal} only includes the 27
publications that were selected for this study and subsequently forms the
basis for answering the research questions.

While DSS for fisheries and aquaculture have a long history going back
to the early 1980's, e.g. Scuse et al.\cite{scuse1983information},
the literature surrounding them is sparse. Figure~\ref{fig:journal}
shows several periods during the last 25 years in which  the systematic mapping
study did not identify any relevant literature published with regards to the
criteria previously defined. These periods include
the years of 1996, 1997, 2003, 2004, 2010-2012 and
2015. Figure~\ref{fig:journal} also shows that the topic was at its
most popular during the early 2000's, i.e. 2000-2002. The subject also
saw increased popularity in the period of 2007-2009, after which the
publication rate fell drastically, as evidenced by the fact that the
last five years have seen only one relevant published paper in the
form of (SID-27) in 2013. These periods of elevated interest are
consistent with the result presented in Arnott et
al.~\cite{arnott2014critical}, which also notes an increase in
publications in these periods.

The declining DSS publishing trend of the last 5 years is not unique to the
fishery and aquaculture disciplines as pointed out by
~\cite{arnott2014critical}, who notes an overall decline in the number of DSS
related publications since the early 1990's. Arnott et
al.~\cite{arnott2014critical} speculates that the decline in DSS publications
might stabilize in the coming years as DSS reaches a more balanced position
within the domain of information systems, noting that the declining use of DSS
might be due to the adoption of other models like the technology acceptance
model. While the publishing trend is currently declining,
Figure~\ref{fig:journal} shows that decision support systems within fisheries
and aquaculture is still being researched, but to a lesser degree than in
previous years.

According to Mardle and Pascoe~\cite{mardle1999review}(SID-6), there
are few publications on multi-criteria decision making within the
field of fishery management compared to other fields like forestry,
agriculture, and, finance. SID-6 postulates that in general, the
more publications that appear for a given topic, the more research
is stimulated, and, thus further publications are generated. The lack
of publishing results could then according to Mardle and Pascoe
in SID-6 result in an unwillingness to adopt the MCDM technique among
the decision makers. Simultaneously, SID-6 conclude that
multi-criteria decision making can play an important role in the
development of fisheries management policy. The following sections
provide a discussion of how each research question was addressed in
the mapping study. Results from the mapping study will be presented
for each RQ, followed by a discussion of their implications.

\subsubsection{(RQ-1) What decision support systems for fisheries and
  aquaculture exists?}
\label{sec:rq1}
The DSSs found during the review part of the mapping study can
coarsely be divided into two categories in accordance with RQ1. The
results of this mapping are presented in Table~\ref{tab:separation}.
\tabulinesep=1.2mm
\begin{center}
  \begin{table}[H] 
  \centering    
  \begin{tabu} to 1\columnwidth {  X[1,l]  X[2,l]  }
  \hline
  Category & Study ID \\ \hline 
  Fishery & 3, 4, 5, 10, 11, 12, 13, 14, 16, 17, 18, 19, 
  20, 22, 23, 25, 26, 27 \\ 
  Aquaculture & 1, 2, 6, 7, 8, 9, 14, 15, 21, 24 \\ \hline
  \end{tabu}
  \caption{Studies separated into fishery and aquaculture categories
    by topic.}
  \label{tab:separation}
\end{table}
\end{center}

Table~\ref{tab:separation} show DSS research to be more heavily
focused around fisheries rather than aquaculture; however both fields
are well anchored in research and case studies alike.

Noticeably only a single decision support system from
Table~\ref{tab:separation} integrates both fishery and aquaculture
requirements in their decision making process, though it should be
noted that they are logically separated in the respected
system. Except for SID-14 all systems identified in the study focused
on either fishery or aquaculture, and the general trend indicates that
they should be managed as two distinct classes.

A further classification using the type of DSS described in the papers
has been performed, and is shown in Figure~\ref{tab:dsstypes}. Column
one of Table~\ref{tab:dsstypes} shows the classification of DSS within
fisheries, while column two shows the same classification for
aquaculture. As can be seen from Table~\ref{tab:dsstypes}, the
majority of the fishery DSS systems utilize either a model-driven
architecture or neural networks. It seems to be no data driven and
significantly few of the remaining types found in this study. This
indicates that model-driven- or neural network DSS is the most fitting
and applicable methodologies for the fishery DSS domain.
\begin{center}
  \begin{table}[H] 
  \centering    
  \begin{tabu} to 1\columnwidth {  X[1.4,l]  X[1,l]  X[1,l]  }
  \hline
  DSS type & Fishery studies & Aquaculture studies\\
  \hline
  Model driven DSS & 7 & 4\\  
  Geographical DSS & 1 & 2\\  
  Multi-Criteria DSS & 4 & 4\\ 
  Grid Based DSS & 1 & 0\\  
  Knowledge Based DSS & 1 & 1\\
  Multi-Objective DSS & 1 & 0\\ 
  Data-Driven DSS & 0 & 1\\  
  Neural Network DSS & 3 & 0\\  \hline
  \end{tabu}
  
  \caption{DSS types grouped by Fishery and Aquaculture.}
  \label{tab:dsstypes}
\end{table}
\end{center}

From Table~\ref{tab:dsstypes} we can see that model-driven
architectures are also popular within aquaculture DSS, however we note
that multi-criteria systems are significantly more utilized within
aquaculture. Table~\ref{tab:dsstypes} shows that model-driven systems
remain the overall most popular design, but also suggest that the
aquaculture and fishery domains contain distinct requirements, which
are best solved by differing methods. Also one can observe that there is very
little overlap, the sum of types of DSSs in Table~\ref{tab:dsstypes} is 30 and
the number of studies is 27 so only a very few studies actually inhabits more
than one type. There is no reason why a DSS using Geographical models cannot
also be using neural networks as well. The most likely reasons is research
method (focusing on measuring effect of one type of DSS rather than multiple)
and effort (implementing more than one is more resources demanding)

\subsubsection{(RQ-2) What are the most investigated fishery
and aquaculture DSS topics and how have these changed over time.}
\label{sec:rq2}
In addition to classifying the result set based on their type, as seen
in Table~\ref{tab:dsstypes}, the papers were also classified by their
problem domains. The results of this classification are presented in
Table~\ref{tab:topics}, which shows what papers corresponds to what
topic, and Figure~\ref{fig:timeline} in Appendix B which shows how the popularity of the
different topics has changed over time. It should be noted that
individual papers can correspond to multiple topics in both
Table~\ref{tab:topics} and the table in Appendix B.
\begin{center}
  \begin{table}[H] 
  \centering    
  \begin{tabu} to 1\columnwidth {  X[1.4,l]  X[2,l]  }
  \hline
Associated DSS topic &
Study ID (SID) \\ \hline
Fish health and disease management &
26 \\
Scheduling and planning &
14, 16, 25 \\
Harvest regulations &
4, 13, 19, 22, 25 \\ 
Sustainability &
4, 11, 13, 17, 19, 22, 25, 27 \\ 
Catch optimization &
12, 13, 17, 19, 20, 22, 23, 25 \\ 
Management decisions & 
1, 2, 3, 6, 7, 8, 11, 13, 14, 16, 22, 23, 27 \\ \hline
  \end{tabu}
  
  \caption{DSS systems grouped by topics.}
  \label{tab:topics}
\end{table}
\end{center}

Decision support systems for fisheries and aquaculture are very
complex tasks as can be seen in e.g. Ernst et
al.~\cite{ernst2000aquafarm}, and resolving all parameters
influencing decisions which can be mapped to real world scenarios is
difficult. This study shows that most fishery- and aquaculture DSS
only takes a small subset of criteria into consideration and thus
apply only to a single species or a small subset of related species as
shown in Table~\ref{tab:topics}. While there are some exceptions to
the case where papers cover multiple topics, e.g. SID-13, 19, and 25,
it is worth noting that some of the topics covered are often closely
related, as is the case with harvest regulations and
sustainability. As such, it is natural that papers will often go into
related topics as can be seen in Table~\ref{tab:selectedstudies} in
Appendix B, especially when using the categorization of topics that
are defined in Table~\ref{tab:topics}. Utilizing the categorization of
topics that we have chosen, we found three papers (SID-13, 22 and 25)
represented in five different categories, whereas the most common is
that a paper is represented in only one or two topics.

Regression analysis performed in Microsoft Excel did not show any
noticeable trends regarding the popularity of different topics, and
how these topics changed over time, a result most likely due to the
small data-set. We reiterate the fact that there are two periods of
elevated research in published research i.e. around 2001 and 2009;
however these periods are limited in the amount of literature
published.  This indicates that papers are usually occurring in bulks
during a narrow time period, and general purpose DSS as the most
recurring theme.

From the table in Appendix B we may infer sustainability and catch
optimizations as the most researched form of DSS topics during the
most recent years. Considering the increased concern in the most
recent years of the global over-fishing problem, it reflects that
fishery- and aquaculture DSS might be indicative of political
tendencies, especially when coupled with monetary
interests. Therefore, we are not surprised to see efficiency and
sustainability as popular topics, as these are universal interests
that persist across all industries. 

Lastly we can see that the analysis of the studies shown in
Table~\ref{tab:topics} shows more overlap than the analysis presented in
Table~\ref{tab:dsstypes}. Thus more studies claim to be affecting more topics
than they are using different DSS methods/types. Given the hypothesis explaining 
Table~\ref{tab:dsstypes} very low degree of overlap - The higher degree of overlap
in Table~\ref{tab:topics} would suggest that it is less resource intensive to
test the DSS in another topic and/or that it is less confounding
methodologically analyzing the effects of a single DSS type on different topics
rather than analyzing different DSS types on one topic.

\subsubsection{(RQ-3) Does there exist any DSSs for fisheries
  providing real-time analytics?}
\label{sec:rq3}
Given the low number of related papers found during the search phase, we found
only one paper (SID-19) to fulfill the requirements to be a complete real time
fishery DSS, while SID-16, 17 and 23 contains real-time components. We
identified sustainability as an important topic in RQ-2 reflecting the growing
problem of global over-fishing. As a consequence management decisions should
account for sustainable fishing and water pollution in their decision making
process. To this end, Caddy and Mahon\cite{caddy1995reference} outlines
reference points for fishery management which provides quantitative indicators
for system objectives. It states that developing reference points to make
quality decisions should be substantiated not only in historical data, but also
real time data to best solve their objective. Thus facilitating the usage of
real time-data is important for fishery- and aquaculture DSS involving fishing
regulations or sustainability. Despite the limited research on this topic,
SID-19 shows that real-time data and analytics is possible and has been done
previously, although for a narrow area.

\subsubsection{(RQ-4) Do DSSes in Fisheries and Aquaculture build their models
  using machine learning done on captured and grounded data?}
\label{sec:rq4}

An important quality for a DSS is how well it represents the true situation of
the decision problem. This quality can be threatened by using out of date
models or out of date input data (e.g. simulation or model parameters). In
Table~\ref{tab:mltable} we have summarized usage of machine learning and what
data the machine learning was trained on. In this context we define machine
learning as creating models of phenomenon (e.g. for prediction as in SID-17)
automatically based on data. This is in opposition to more static AI methods
such as rule based expert systems (e.g. see the traffic light system of Hargrave
in SID-15), which is not considered machine learning in this table.

\begin{table}
\begin{center}
  \caption[Studies using machine learning.]{Studies using machine learning.}\label{tab:mltable}
\begin{tabu} to 1\linewidth {X[l,4] X[l,6] }
  \hline
  Sub-domain & SID \\
  \hline
  Fisheries & 12, 16, 17, 20, 25 \\
  Aquaculture & 26  \\ 
  \hline
\end{tabu}
\end{center}
\end{table}

We can see from Table~\ref{tab:mltable} that there is only 6 publications that
apply any kind of machine learning as a part of their main contribution. More
profound in Aquaculture only 1 study (8,33\%) uses machine learning. In
comparison Fisheries has 5 publications (31,5\%) using machine learning as a
part of their main contribution.

In Fisheries and Aquaculture data should be available through operational
requirements (e.g. video monitoring in aquaculture and catch logging in
fisheries). It is an obvious disadvantage for DSS scientists to not draw on
these resources for model creation and prediction. This should be a area of
focus in future DSS research within this domain.

\section{Discussion}
\label{sec:discussion}

This paper presented a systematic mapping study conducted in order to create a
solid theoretical foundation for research on DSSs in fisheries and aquaculture
by looking at existing research. To our knowledge this is the first mapping
study within the field of fishery and aquaculture DSS. The mapping study
considered a time period of 25 years from 1990 to the first half of 2015, and
resulted in 27 papers that satisfy both the research questions and the
inclusion- and exclusion criteria. The results has been validated by the two
authors that did not take part in the initial screening (see
Section~\ref{validity}) that performed a random test to reduce the possibility
of any relevant papers being missed when performing the search phase. The
validation did not reveal any absence of significant papers, and therefore we
believe these 27 papers to accurately describe the current state of the research
and work in fishery and aquaculture DSS.

Noticeably throughout the study, very few of the papers found contained rigorous
empirical evaluation (see Table~\ref{tab:validationtable}), however most
contained an overview of the system and an outline of their methodology. Some
studies used validation data sets, others such as SID-15 actually validated
their results against experts and actual decisions. Some studies claims
validation but with so short descriptions that the reader is left uncertain with
regards to the validation process.

\begin{table}[h!]
\begin{center}
  \caption[Validation of studies.]{Validation of studies.}\label{tab:validationtable}
\begin{tabu} to 1\linewidth {X[l,4] X[l,6] }
  \hline
  Type of validation & SID \\
  \hline
  Validated & 15, 19, 26 \\
  Validation data-set & 17, 18, 20, 25, 27 \\ 
  Weakly described validation & 12, 21  \\
  \hline
\end{tabu}
\end{center}
\end{table}

Although the resulting set of papers do not provide a unified view of DSS
practice, they offer a broad picture and experience on the problem domain. The
resulting set of papers could provide those looking to create new DSS within the
fishery and aquaculture domains with some helpful insights regarding what
methodology is best suited in the design of the DSS depending on the target
domain, and an indication of the possibilities within DSSs.

Looking at the research found throughout the mapping study it can be noted that
none of the resulting set of papers adequately addresses how to create a
decision support system for multiple stakeholders, with multiple criteria with
cross cutting concerns for both fisheries and aquaculture. The results of the
mapping study provide the reader with several useful and relevant sources of
information that could be helpful in the design or research on DSSs in fisheries
and aquaculture.

Fishery- and Aquaculture decision support systems aid management as a high level
of abstraction on the basis of large quantities of data. Using predictive
models, the systems make a qualified guess to increase the probability of an
optimal outcome. Designing a system which uses constant up-to-date data for
running simulations/applying these data to models poses several difficulties.
First and foremost the system is required to keep continuous connections to the
various data sources to manage the data; furthermore the system is responsible
to complete its analysis within a given deadline. These responsibilities often
conflict as no downtime can be expected to complete the deadline
responsibilities of the system, and updating data can cause analytics processes
to be invalidated during its execution.

We believe most DSS for fisheries and aquaculture does not need to be real-time
systems to aid management, and the increased value is often given low priority
because of an increased cost, both complexity and monetary.

The mapping study highlights the lack of fishery and aquaculture decision
support systems encompassing a multitude of factors for a large geographical
area and the species therein. Combining these factors with real-time analytics
has not been attempted, most likely due to the complexity of the system (e.g.
\cite{ernst2000aquafarm}).

\subsection{Threats to validity}
\label{sec:threats}
The primary threats to this study can be identified as whether our approach
adequately addresses the principal research questions. The research questions
require an assessment of the selected data sources to determine whether we
identified all relevant publications and whether our initial classification of
the problem domain is denoted correctly for analysis. As there did not exist any
previous systematic mapping study on the search terms may miss some terms or
synonyms thus missing some publications in the result-set.

The first threat is influenced by both our search criteria, the scope of the
search, and the search terms used which are again limited by the search engines
capabilities. Our primary method for avoiding these pitfalls has been to employ
our third party reviewers to perform a random test on the search queries, the
problem domain etc. to find papers we missed in our systematic mapping study.
The third party did not find any significant papers missed in this study;
therefore we argue that we did manage to provide the most relevant documents
published in scientific journals and computer scrience literature. However,
unpublished studies have not been considered for this study, and it is therefore
possible that we might have missed relevant studies, but overall we presume the
study adequately addresses the principal research questions. There exists no
former mapping study on this particular domain that the authors could find.
Because the topics addressed in this study are selective, there exists little
evidence in either direction that we have omitted a major topic that provides
substantial empirical evaluation which is in direct relevance to this paper.
However it must be noted that we, (the authors), are all software engineers
primarily and not researchers in academia, and might therefore be biased in our
selection and our analysis, although we have worked to the best of our abilities
to avoid this.

\section{Conclusion}
\label{sec:conclusion}
The finding of our systematic mapping study has implications for those
looking to design a multi-criteria decision support system for
fisheries and aquaculture. Our findings show that the majority of
papers evaluating and documenting their methodology did so with few
criteria in mind. The systematic mapping study has identified that
very little research has been made into DSS in fisheries and there is
a lack of empirical evaluation of these systems. Another concern
raised by the study indicates that there exists few to none well
documented multi-criteria DSS systems for both fisheries and
aquaculture. However the analysis has shown that many components for
smaller DSS systems has been well documented, and the result from this
study can work as a foundation for further research and development,
especially coupled with previous research and knowledge on managing
heterogeneous data systems.

\section{Acknowledgment}
\label{sec:acknowledgement}
The basis for this work was done in the eSushi project (RCN grant
number 245951). Parts of the work were also supported by the EXPOSED
SFI project (RCN grant number 237790).

\bibliographystyle{elsarticle-num-names}
\bibliography{dss}

\begin{thebibliography}{1}
\expandafter\ifx\csname url\endcsname\relax
  \def\url#1{\texttt{#1}}\fi
\expandafter\ifx\csname urlprefix\endcsname\relax\def\urlprefix{URL }\fi
\expandafter\ifx\csname href\endcsname\relax
  \def\href#1#2{#2} \def\path#1{#1}\fi

\bibitem{Einstein}
A.~Einstein, {Zur Elektrodynamik bewegter K{\"o}rper}. ({German}) [{On} the
  electrodynamics of moving bodies], Annalen der Physik 322~(10) (1905)
  891--921.
\newblock \href {http://dx.doi.org/http://dx.doi.org/10.1002/andp.19053221004}
  {\path{doi:http://dx.doi.org/10.1002/andp.19053221004}}.

\end{thebibliography}


\begin{thebibliography}{16}
\providecommand{\natexlab}[1]{#1}
\providecommand{\url}[1]{\texttt{#1}}
\providecommand{\urlprefix}{URL }
\expandafter\ifx\csname urlstyle\endcsname\relax
  \providecommand{\doi}[1]{doi:\discretionary{}{}{}#1}\else
  \providecommand{\doi}[1]{doi:\discretionary{}{}{}\begingroup
  \urlstyle{rm}\url{#1}\endgroup}\fi
\providecommand{\bibinfo}[2]{#2}

\bibitem[{Lannan(1993)}]{lannan1993user}
\bibinfo{author}{J.~Lannan}, \bibinfo{title}{User's Guide to PONDCLASS:
  Guidelines for Fertilizing Aquaculture Ponds}, \bibinfo{journal}{Pond
  Dynamics/Aquaculture CRSP, Oregon State University, Corvallis, Oregon} .

\bibitem[{Power(2008)}]{power2008decision}
\bibinfo{author}{D.~J. Power}, \bibinfo{title}{Decision support systems: a
  historical overview}, in: \bibinfo{booktitle}{Handbook on Decision Support
  Systems 1}, \bibinfo{publisher}{Springer}, \bibinfo{pages}{121--140},
  \bibinfo{year}{2008}.

\bibitem[{Garg et~al.(2005)Garg, Adhikari, McDonald, Rosas-Arellano, Devereaux,
  Beyene, Sam, and Haynes}]{garg2005effects}
\bibinfo{author}{A.~X. Garg}, \bibinfo{author}{N.~K. Adhikari},
  \bibinfo{author}{H.~McDonald}, \bibinfo{author}{M.~P. Rosas-Arellano},
  \bibinfo{author}{P.~Devereaux}, \bibinfo{author}{J.~Beyene},
  \bibinfo{author}{J.~Sam}, \bibinfo{author}{R.~B. Haynes},
  \bibinfo{title}{Effects of computerized clinical decision support systems on
  practitioner performance and patient outcomes: a systematic review},
  \bibinfo{journal}{Jama} \bibinfo{volume}{293}~(\bibinfo{number}{10})
  (\bibinfo{year}{2005}) \bibinfo{pages}{1223--1238}.

\bibitem[{Hunt et~al.(1998)Hunt, Haynes, Hanna, and Smith}]{hunt1998effects}
\bibinfo{author}{D.~L. Hunt}, \bibinfo{author}{R.~B. Haynes},
  \bibinfo{author}{S.~E. Hanna}, \bibinfo{author}{K.~Smith},
  \bibinfo{title}{Effects of computer-based clinical decision support systems
  on physician performance and patient outcomes: a systematic review},
  \bibinfo{journal}{Jama} \bibinfo{volume}{280}~(\bibinfo{number}{15})
  (\bibinfo{year}{1998}) \bibinfo{pages}{1339--1346}.

\bibitem[{Kawamoto et~al.(2005)Kawamoto, Houlihan, Balas, and
  Lobach}]{kawamoto2005improving}
\bibinfo{author}{K.~Kawamoto}, \bibinfo{author}{C.~A. Houlihan},
  \bibinfo{author}{E.~A. Balas}, \bibinfo{author}{D.~F. Lobach},
  \bibinfo{title}{Improving clinical practice using clinical decision support
  systems: a systematic review of trials to identify features critical to
  success}, \bibinfo{journal}{Bmj}
  \bibinfo{volume}{330}~(\bibinfo{number}{7494}) (\bibinfo{year}{2005})
  \bibinfo{pages}{765}.

\bibitem[{Crossland et~al.(1995)Crossland, Wynne, and
  Perkins}]{crossland1995spatial}
\bibinfo{author}{M.~D. Crossland}, \bibinfo{author}{B.~E. Wynne},
  \bibinfo{author}{W.~C. Perkins}, \bibinfo{title}{Spatial decision support
  systems: An overview of technology and a test of efficacy},
  \bibinfo{journal}{Decision support systems}
  \bibinfo{volume}{14}~(\bibinfo{number}{3}) (\bibinfo{year}{1995})
  \bibinfo{pages}{219--235}.

\bibitem[{Malczewski(2006)}]{malczewski2006gis}
\bibinfo{author}{J.~Malczewski}, \bibinfo{title}{GIS-based multicriteria
  decision analysis: a survey of the literature},
  \bibinfo{journal}{International Journal of Geographical Information Science}
  \bibinfo{volume}{20}~(\bibinfo{number}{7}) (\bibinfo{year}{2006})
  \bibinfo{pages}{703--726}.

\bibitem[{Kitchenham and Charters(2007)}]{kitchenham2007}
\bibinfo{author}{B.~A. Kitchenham}, \bibinfo{author}{S.~Charters},
  \bibinfo{title}{Guidelines for performing systematic literature reviews in
  software engineering}, \bibinfo{type}{Tech. Rep.},
  \bibinfo{institution}{Keele University}, \bibinfo{year}{2007}.

\bibitem[{Budgen et~al.(2008)Budgen, Turner, Brereton, and
  Kitchenham}]{budgen2008using}
\bibinfo{author}{D.~Budgen}, \bibinfo{author}{M.~Turner},
  \bibinfo{author}{P.~Brereton}, \bibinfo{author}{B.~Kitchenham},
  \bibinfo{title}{Using mapping studies in software engineering}, in:
  \bibinfo{booktitle}{Proceedings of PPIG}, vol.~\bibinfo{volume}{8},
  \bibinfo{organization}{Lancaster University}, \bibinfo{pages}{195--204},
  \bibinfo{year}{2008}.

\bibitem[{Petticrew and Roberts(2008)}]{petticrew2008systematic}
\bibinfo{author}{M.~Petticrew}, \bibinfo{author}{H.~Roberts},
  \bibinfo{title}{Systematic reviews in the social sciences: A practical
  guide}, \bibinfo{publisher}{John Wiley \& Sons}, \bibinfo{year}{2008}.

\bibitem[{Kitchenham et~al.(2007)Kitchenham, Mendes, and
  Travassos}]{kitchenham2007cross}
\bibinfo{author}{B.~A. Kitchenham}, \bibinfo{author}{E.~Mendes},
  \bibinfo{author}{G.~H. Travassos}, \bibinfo{title}{Cross versus
  within-company cost estimation studies: A systematic review},
  \bibinfo{journal}{Software Engineering, IEEE Transactions on}
  \bibinfo{volume}{33}~(\bibinfo{number}{5}) (\bibinfo{year}{2007})
  \bibinfo{pages}{316--329}.

\bibitem[{Scuse and Arnason(1983)}]{scuse1983information}
\bibinfo{author}{D.~H. Scuse}, \bibinfo{author}{A.~N. Arnason},
  \bibinfo{title}{INFORMATION MANIPULATION IN BIOLOGICAL DECISION-SUPPORT
  SYSTEMS}, in: \bibinfo{booktitle}{Proceedings of the Sixteenth Hawaii
  International Conference on System Sciences, 1983}, vol.~\bibinfo{volume}{1},
  \bibinfo{organization}{Western Periodicals Company}, \bibinfo{pages}{377},
  \bibinfo{year}{1983}.

\bibitem[{Arnott and Pervan(2014)}]{arnott2014critical}
\bibinfo{author}{D.~Arnott}, \bibinfo{author}{G.~Pervan}, \bibinfo{title}{A
  critical analysis of decision support systems research revisited: the rise of
  design science}, \bibinfo{journal}{Journal of Information Technology}
  \bibinfo{volume}{29}~(\bibinfo{number}{4}) (\bibinfo{year}{2014})
  \bibinfo{pages}{269--293}.

\bibitem[{Mardle and Pascoe(1999)}]{mardle1999review}
\bibinfo{author}{S.~Mardle}, \bibinfo{author}{S.~Pascoe}, \bibinfo{title}{A
  review of applications of multiple-criteria decision-making techniques to
  fisheries}, \bibinfo{journal}{Marine Resource Economics}
  (\bibinfo{year}{1999}) \bibinfo{pages}{41--63}.

\bibitem[{Ernst et~al.(2000)Ernst, Bolte, and Nath}]{ernst2000aquafarm}
\bibinfo{author}{D.~H. Ernst}, \bibinfo{author}{J.~P. Bolte},
  \bibinfo{author}{S.~S. Nath}, \bibinfo{title}{AquaFarm: simulation and
  decision support for aquaculture facility design and management planning},
  \bibinfo{journal}{Aquacultural Engineering}
  \bibinfo{volume}{23}~(\bibinfo{number}{1}) (\bibinfo{year}{2000})
  \bibinfo{pages}{121--179}.

\bibitem[{Caddy and Mahon(1995)}]{caddy1995reference}
\bibinfo{author}{J.~F. Caddy}, \bibinfo{author}{R.~Mahon},
  \bibinfo{title}{Reference points for fisheries management}, vol.
  \bibinfo{volume}{374}, \bibinfo{publisher}{Food and Agriculture Organization
  of the United Nations Rome}, \bibinfo{year}{1995}.

\end{thebibliography}

\appendix
\onecolumn

\section{Selected studies}
\label{sec:selectedstudies}

  \begin{longtabu} to 1\textwidth { | X[1,l] | X[10,l] | }

\caption[Table with selected studies.]{The selected studies.}\label{tab:selectedstudies}\\ 

\hline
Study ID (SID) & Reference \\ 
\endfirsthead
\caption[]{(Cont.) The selected studies.}\\  
\hline
Study ID (SID) & Reference\\   \hline
\endhead
\hline
\endfoot

\hline
1 &
G. Bourke, F. Stagnitti and B. Mitchell, "A decision support system
for aquaculture research and management". In: Aquacultural
Enginerring(1993)\\ \hline

2 &
W. Silvert,"A decision support system for regulating finfish
aquaculture". In: Ecological Modelling(1994)\\ \hline

3 &
I. Renaud and S. Yacout, "Decision support system for quality
assurance programs in the fish and seafood processing industry". In:
Computers \& Industrial Engineering(1995)\\ \hline

4 &
J. Korrûbel, S. Bloomer, K. Cochrane, L. Hutchings and J. Field,
"Forecasting in South African pelagic fisheries management: the use of
expert and decision support systems". In: South African Journal of
Marine Science (1998)\\ \hline

5 &
S. Mardle and S. Pascoe, "A review of applications of
multiple-criteria decision-making techniques to fisheries". In: Marine
Resource Economics(1999)\\ \hline

6 &
J. Bolte, S. Nath and D. Ernst, "Development of decision support tools
for aquaculture: the POND experience". In: Aquacultural Engineering
(2000)\\ \hline

7 &
O. F. El-Gayar and P. Leung, "ADDSS: A tool for regional aquaculture
development". In: Aquacultural Engineering (2000)\\ \hline

8 &
D. H. Ernst, J. P. Bolte and S. S .Nath, "AquaFarm: Simulation and
decision support for aquaculture facility design and management
planning". In: Aquacultural Engineering (2000)\\ \hline

9 &
S. S. Nath, J. P. Bolte, L. G. Ross and J. Aguilar-Manjarrez,
"Applications of geographical information systems (GIS) for spatial
decision support in aquaculture". In: Aquacultural Engineering(2000)\\
\hline

10 &
S. R. Coppola and D. Crosetti, "Decision-support Systems for Fisheries
the "ITAFISH" Case Study". In: Studies and Reviews - General Fisheries
Commission for the Mediterranean, No. 72 (2001)\\ \hline

11 &
K. Hughey, R. Cullen, A. Memon, G. Kerr and N. Wyatt, "Developing a
Decision Support system to manage fisheries externalities in New
Zealand's Exclusive Economic Zone". In: Modeling and Economic Theory
(IIFET 2000) (2001)\\ \hline

12 &
S. Mackinson, "Integrating local and scientific knowledge: an example
in fisheries science". In: Environmental Management (2001)\\ \hline

13 &
M. Pan, P. S. Leung and S. G. Pooley, "A decision support model for
fisheries management in Hawaii: a multilevel and multiobjective
programming approach". In: North American Journal of Fisheries
Management (2001)\\ \hline

14 &
R. Pastres, C. Solidoro, G. Cossarini, D. M. Canu and C. Dejak,
"Managing the rearing of Tapesphilippinarum in the lagoon of Venice: a
decision support system". In: Ecological Modelling (2001)\\ \hline

15 &
B. T. Hargrave , "A traffic light decision system for marine finfish
aquaculture siting". In: Ocean and Coastal Management (2002)\\ \hline

16 &
Z. Kemp and G. Meaden, "Visualization for fisheries management from a
spatiotemporal perspective". In: ICES Journal of Marine Science:
Journal duConsell (2002)\\ \hline

17 &
A. Iglesias, B. Arcay, A. Rodriguez and M. Cotos,"A support system for
fisheries based on neural networks". In: 1st International Workshop on
Artificial Neural Networks and Intelligent Information Processing,
ANNIIP 2005- In Conjunction with ICINCO2005 (2005)\\ \hline

18 &
T. Koutroumanidis, L. Iliadis and G. K. Sylaios, "Time-series modeling of fishery landings using
ARIMA models and Fuzzy Expected Intervals software". In: Environmental
Modelling \& Software (2006)\\ \hline
19 & 
N. Carrick and B. Ostendorf, "Development of a spatial Decision
Support System (DSS) for the Spencer Gulf penaeid prawn fishery, South
Australia". In: Environmental Modelling \& Software (2007)\\ \hline

20 &
A. Iglesias, C. Dafonte, B. Arcay and J. M. Cotos, "Integration of
remote sensing techniques and connectionist models for decision
support in fishing catches". In: Environmental Modelling \& Software
(2007)\\ \hline

21 &
R. Wang, D. Chen and Z. Fu, "AWQEE-DSS: A decision support system for
aquaculture water quality evaluation and early-warning". In: 2006
International Conference on Computation Intelligence  and Sequrity
ICCIAS 2006 (2007)\\ \hline

22 &
F. Azadivar, T. Truong and Y. Jiao, "A decision support system for
fisheries management using operations research and systems science
approach". In: Expert Systems with Applications (2009)\\ \hline

23 &
R. V. Chandran, A. Jeyaram, V. Jayaraman, S. Manoj, K. Rajitha and
C. K. Mukherjee, "Prioritization of satellite-derived potential
fishery grounds: An analytical hierarchical approach-based model using
spatial and non-spatial data". In: International Journal of Remote
Sensing (2009)\\ \hline

24 &
H. Halide, A. Stigebrandt, M. Rehbein and A. McKinnon, "Developing a
decision support system for sustainable cage aquaculture". In:
Environmental Modelling \& Software (2009)\\ \hline

25 &
L. Sun, H. Xiao, D. Yang and S. Li, "Intelligent decision support
system for fisheries management". In: Journal of Computational
Information Systems (2009)\\ \hline

26 &
Z. Xiaoshuan, F. Zetian, C. Wengui, T. Dong and Z. Jian, "Applying
evolutionary prototyping model in developing FIDSS: An intelligent
decision support system for fish disease/health management". In:
Expert Systems with Applications (2009)\\ \hline

27 &
Y. K.Teniwut and Marimin, "Decision support system for increasing
sustainable productivity on fishery agroindustry supply chain". In:
5th International Conference on Advanced Computer Science and
Information Systems, ICACSIS (2013)\\

  \end{longtabu}
  
\section{Timeline of DSS topics}
\label{sec:timeline}

\begin{figure}[H]
  \centering
    \includegraphics[width=\textwidth]{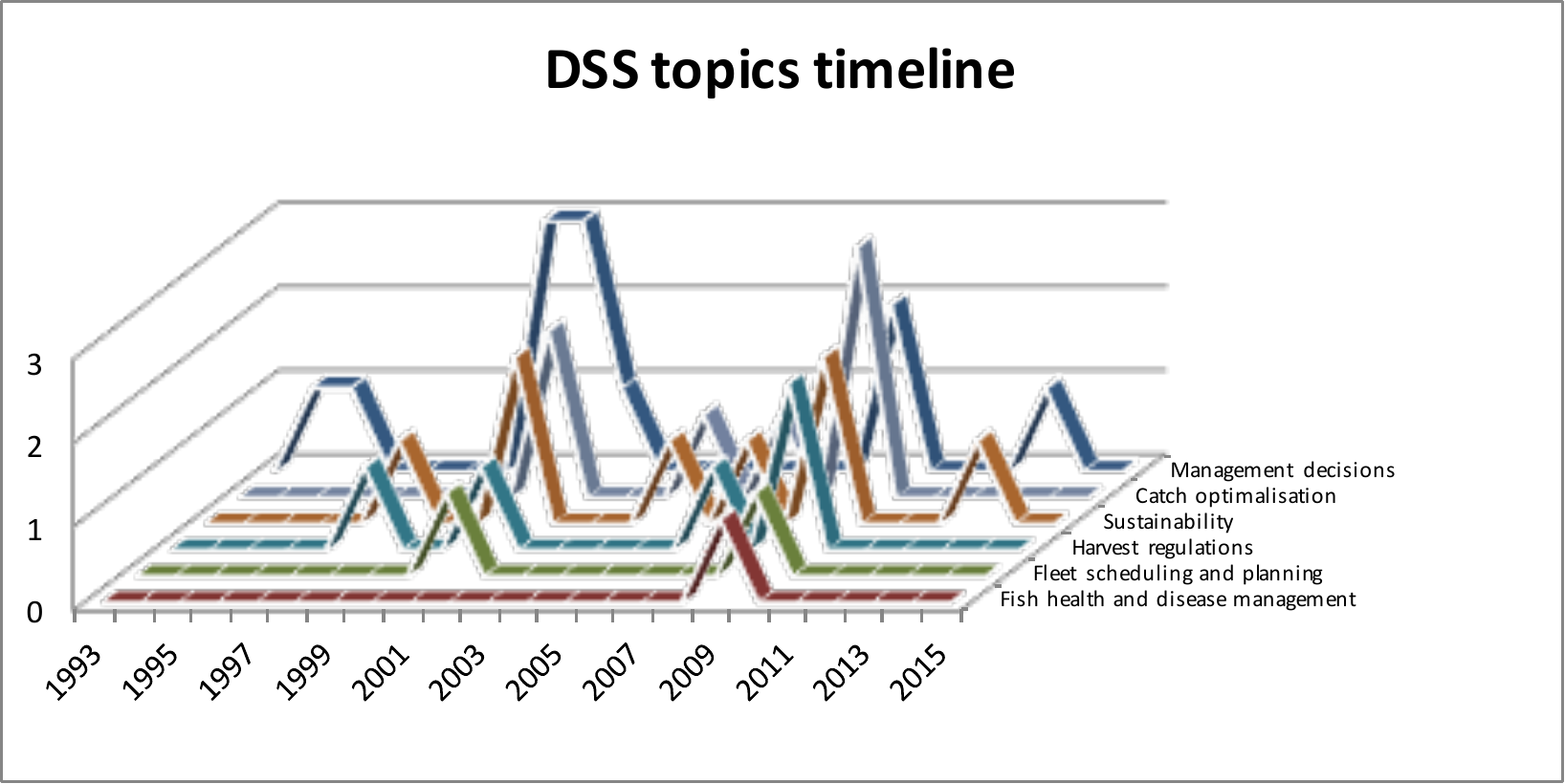}
    \caption{Timeline of DSS topics in selected studies}
  \label{fig:timeline}
\end{figure}

\end{document}